\documentclass{article}

    \usepackage[preprint]{neurips_2025}

\usepackage[utf8]{inputenc} % allow utf-8 input
\usepackage[T1]{fontenc}    % use 8-bit T1 fonts
\usepackage{hyperref}       % hyperlinks
\usepackage{url}            % simple URL typesetting
\usepackage{booktabs}       % professional-quality tables
\usepackage{amsfonts}       % blackboard math symbols
\usepackage{nicefrac}       % compact symbols for 1/2, etc.
\usepackage{microtype}      % microtypography
\usepackage{xcolor}         % colors

\usepackage{enumitem}
\usepackage{comment}
\usepackage{amssymb}

\usepackage{graphicx}
\usepackage{amsmath} 
\usepackage{float}
\usepackage{wrapfig}

\usepackage{multirow}
\usepackage{multicol}
\usepackage{makecell}

\usepackage{microtype}
\usepackage{graphicx}
\usepackage{subfigure}
\usepackage{booktabs} % for professional tables

\usepackage[ruled,vlined]{algorithm2e}
\usepackage{algorithmic}

\usepackage{caption}

\usepackage{amsthm}
\theoremstyle{plain}
\newtheorem{theorem}{Theorem}[section]

\newtheorem{definition}[theorem]{Definition}

\newtheorem{remark}[theorem]{Remark}

\title{
FastCar: \underline{C}ache \underline{A}ttentive \underline{R}eplay for \\ Fast Auto-Regressive Video Generation on the Edge
}

\author{
  Xuan Shen$^{1*}$, \ Weize Ma$^{2}$\thanks{Equal Contribution} \ , \ Yufa Zhou$^{3}$, \ Enhao Tang$^{2}$, \ Yanyue Xie$^1$, \ Zhengang Li$^4$, \\
  \textbf{
  Yifan Gong$^4$, \ Quanyi Wang$^5$,\ Henghui Ding$^6$,\ Yiwei Wang$^7$,
  }
  \\
  \textbf{
  Yanzhi Wang$^1$\thanks{Corresponding Authors} \ , \ Pu Zhao$^{1\dag}$, \ Jun Lin$^{2\dag}$, \ Jiuxiang Gu$^{4\dag}$
  }
  \\
  $^1$Northeastern University, $^2$Nanjing University, $^3$Duke University, 
  \\
  $^4$Adobe Research
  $^5$NUIST, $^6$Fudan University, 
  $^7$UCM
  \\
  \texttt{shen.xu@northeastern.edu} 
}

\newif\ifmodify 
\modifytrue % CARE: comment this line to compile the final version
\ifmodify

\else

\fi

\begin{document}

\maketitle

\begin{abstract}

Auto-regressive (AR) models, initially successful in language generation, have recently shown promise in visual generation tasks due to their superior sampling efficiency.
% compared to diffusion-based methods. 
Unlike image generation, 
% which requires relatively few tokens to render an image, 
video generation requires a substantially larger number of tokens to produce coherent temporal frames, resulting in significant overhead during the decoding phase. 
% Our detailed observations and analysis find specific deep insights for auto-regressive video generation: (i) MLP modules in decode phase dominate the inference latency and (ii) there exists high temporal redundancy in MLP outputs of adjacent frames. 
Our key observations are: (i) MLP modules in the decode phase dominate the inference latency, and (ii) there exists high temporal redundancy in MLP outputs of adjacent frames. 
In this paper, we propose the \textbf{FastCar} framework to accelerate the decode phase for the AR video generation  by exploring the temporal redundancy. The Temporal Attention Score (TAS) is proposed to determine whether to apply the replay strategy (\textit{i.e.}, reusing cached MLP outputs from the previous frame to reduce redundant computations)  with detailed theoretical analysis and justification.     
%Unlike image generation, which requires relatively few tokens to render an image, video generation requires a substantially larger number of tokens to produce coherent temporal frames, resulting in significant overhead during the decoding phase. 
%This challenge is distinct from the long-context overhead of transformer-based models in language tasks or the high-resolution diffusion tasks.
%In this paper, we propose the \textbf{FastCar} framework for the acceleration of the decoding phase for the auto-regressive video generation.
% Since attention accounts for only a minor portion of the generation-time computation, we explore the feedforward layers by investigating temporal redundancy among generated tokens. Guided by the attention between the current token and its spatially aligned counterpart in the previous frame, we replay the feedforward inference by reusing the cached output associated with the aligned token.
%Since attention accounts for only a minor portion of the generation-time computation, we explore the feedforward layers by investigating temporal redundancy between the current token and its spatially aligned counterpart in the previous frame.
%We further demonstrate that the similarity between these tokens correlates with attention scores, which motivates our replay strategy—reusing cached feedforward outputs from the previous frame to reduce redundant computation.
Also, we develop a hardware accelerator on FPGA with Dynamic Resource Scheduling (DRS) based on TAS to enable better resource utilization and faster inference.
Experimental results demonstrate the effectiveness of our method, which outperforms traditional sparse attention approaches with more than 2.1$\times$ decoding speedup and higher energy efficiency on the edge.
Furthermore, by combining FastCar and sparse attention, FastCar can boost the performance of sparse attention with alleviated drifting, demonstrating our unique advantages for high-resolution and long-duration video generation.
%it complements sparse attention methods by alleviating drifting, offering particular value for future high-resolution and long-duration video generation.}
Code: \textcolor{blue}{\url{https://github.com/shawnricecake/fast-car}}

\end{abstract}

\section{Introduction}

Recently, there has been growing interest in extending the Auto-Regressive (AR) framework of Large Language Models (LLMs)~\cite{gpt2, llama1, llama3} to visual generation tasks~\cite{llamagen, emu3, infinity_bytedance, var_neurips, artv, nova, jiao2025flexvar, xie2024progressive, sun2024autoregressive, luo2024open, kondratyuk2023videopoet, wang2024parallelized}.
The works~\cite{llamagen, var_neurips, sun2024autoregressive, infinity_bytedance, wang2024parallelized}  convert  images into tokens, and apply AR models to generate image tokens with next-token prediction.
The  generation quality is surprisingly strong, often rivaling or surpassing diffusion-based methods~\cite{var_neurips, sun2024autoregressive, infinity_bytedance} in perceptual fidelity and semantic coherence.
% Building on the foundation of CLIP~\cite{clip} and early advances in Visual Language Models (VLMs) works~\cite{liu2023llava, liu2023improvedllava}, emerging efforts~\cite{vilau, unified_model_1, unified_model_2, unified_model_3, unified_model_4, unified_model_5, unified_model_6, unified_model_7} have focused on developing unified generative models that jointly handle language and vision modalities, showcasing the applicability and effectiveness of auto-regressive transformers in multimodal generation tasks.

As video becomes a dominant medium across entertainment, communication, \textit{etc.},  synthesizing coherent high-quality videos from minimal inputs presents a compelling research challenge~\cite{survey_1, survey_2, survey_3, survey_4}. 
%that pushes the limits of generative modeling in both spatial and temporal dimensions.
Prior works~\cite{opensora_plan, opensora1, opensora2, hong2022cogvideo, yang2024cogvideox, kong2024hunyuanvideo} leverage Diffusion Transformers (DiT)~\cite{dit} to develop  video generation models with superior generation performance, at the cost of 
substantial computations and massive memory demands~\cite{he2025neighboring_efficient_ar_image, jin2024pyramidal_efficient_dit_video, xu2025xattention_efficient_sparseattn, xi2025sparse_efficient_dit_video_sparseattn, he2024zipar_efficient_image}. 
%due to intricate complexity  and  iterative sampling   of video diffusion.
%While these methods deliver strong generation performance, they are limited by substantial computational inefficiency and high memory demands~\cite{he2025neighboring_efficient_ar_image, jin2024pyramidal_efficient_dit_video, xu2025xattention_efficient_sparseattn, xi2025sparse_efficient_dit_video_sparseattn, he2024zipar_efficient_image}, as generating high-resolution videos requires prolonged inference times and large GPU memory footprints.
These characteristics limit their applications and deployments for resource-constrained environments~\cite{flightvgm, zhu2023mobilevidfactory, kim2025device, lu2024terdit, shen2024lazydit} such as mobile devices or Field-Programmable Gate Array (FPGA) with tight constraints for energy efficiency,   memory, \textit{etc.}

%These characteristics make DiT-based video generation models particularly unsuitable for resource-constrained environments~\cite{flightvgm, zhu2023mobilevidfactory, kim2025device, lu2024terdit, shen2024lazydit} such as mobile devices or Application-Specific Integrated Circuits (ASICs), where low latency, energy efficiency, and strict memory budgets are critical.
% The need of iterative denoising steps and the lack of streaming-friendly computation further hinder their applicability~\cite{shen2024lazydit, wu2024SnapGenv, li2023snapfusion} in latency-sensitive or on-device video generation scenarios.

Motivated by the scalability and fast decoding capabilities of AR transformer-based frameworks in generative tasks, an increasing number of works~\cite{vilau, nova, artv, xie2024progressive, kondratyuk2023videopoet} have adopted AR frameworks for video generation tasks.
To further improve its efficiency, model compression strategies (such as pruning and quantization~\cite{lin2023awq, ma2023llmpruner, shen2024agile, shen2025numrical, xiao2023smoothquant})  and spatial redundancy optimizations (such as sparse attention~\cite{xiao2023streamingllm, rehg2024kvcompress, hooper2024kvquant, liu2024minicache, ge2024model, li2024snapkv} and  efficient sampling methods~\cite{speculative_1, speculative_2, speculative_3, speculative_4, speculative_5, teng2024accelerating_efficient_image, he2025neighboring_efficient_ar_image, he2024zipar_efficient_image}) are investigated.  However, the inherent temporal redundancy specifically introduced by videos with multiple sequential frames remains largely unexplored in AR video generation. 

\begin{figure*}[t]
  \centering
  \includegraphics[width=1.0\linewidth]{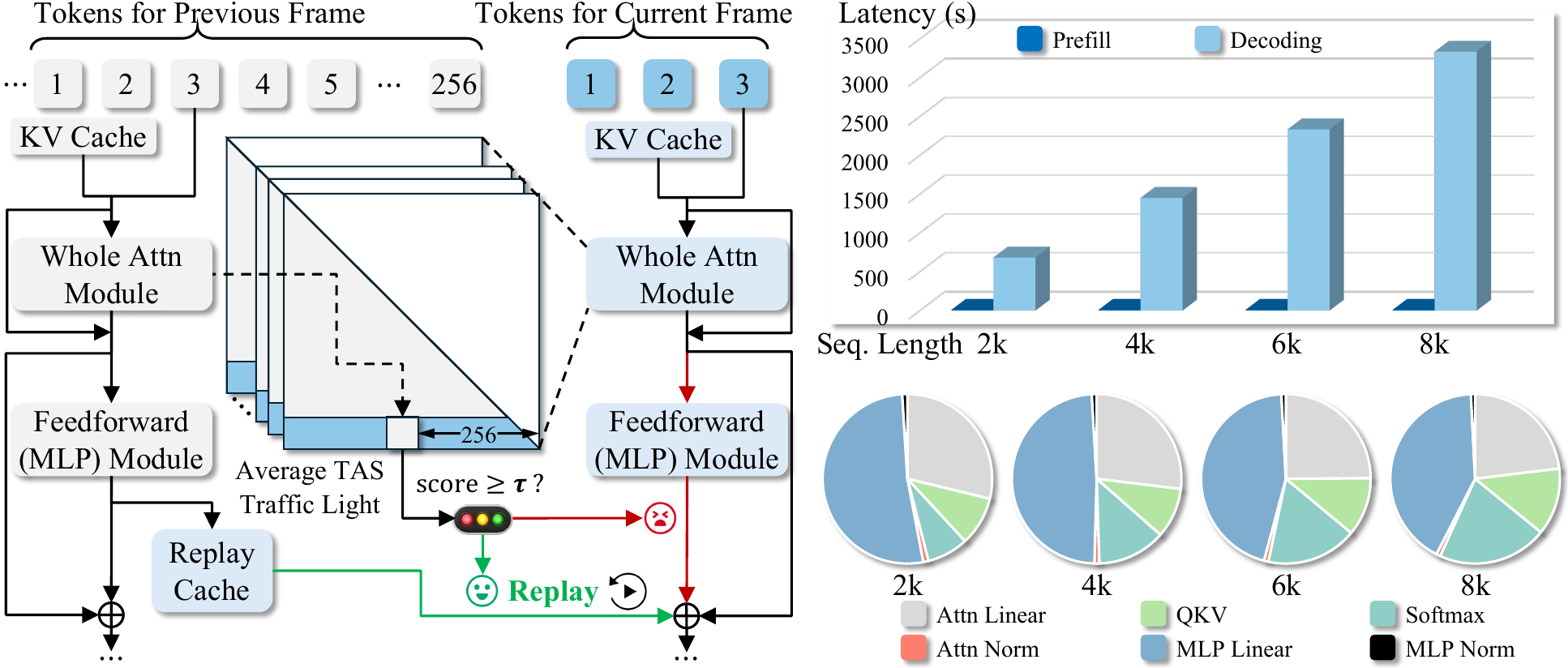}
  \vspace{-4mm}
  \caption{
  % \textbf{Left}: Pipeline of the proposed FastCar framework. We replay the cache from previous frame to skip the computations for the MLP modules in the decoding phase. Replay is triggered when the average TAS exceeds a predefined threshold $\tau$, indicating high similarity across frames.
  % \textbf{Right Top}: Latency cost of both prefill and decoding phases for 2k, 4k, 6k, and 8k sequence length.
  % \textbf{Right Bottom}: Detailed latency cost of the decoding phase for different sequence length.
  \textbf{Left}: FastCar framework. We replay the cache from the previous frame to skip the computations for  MLP in  decoding. Replay is triggered when the average TAS exceeds a predefined threshold $\tau$.
  \textbf{Right Top}: Latency cost of both prefill and decode phases for different sequence lengths.
  \textbf{Right Bottom}: Detailed latency cost of the decode phase for different sequence lengths.
  }
  \label{fig:whole_pipeline}
  \vspace{-2mm}
\end{figure*}

\textbf{Specific Deep Insights.}
To explore the redundancy for superior efficiency, we first perform a detailed latency profiling and a similarity analysis between different frames. As shown in the right of Figure~\ref{fig:whole_pipeline}, we identify that the MLP modules (rather than the attention modules) in the decode phase dominate the inference latency. Meanwhile, according to Figure~\ref{fig:similarity_adjacent_frames}, the outputs of  adjacent frames for the same MLP module exhibit relatively high resemblance/similarity, indicating high temporal redundancy.  

\textbf{Framework with Theoretical Justification.}
Based on the deep insights specific for AR video generation, we propose \textbf{FastCar} for efficiency optimization. We propose the \textit{Temporal Attention Score (TAS)} to determine whether to skip the computations of the MLP modules. If skipped, the cached outputs from the previous frame are directly reused as current outputs (similar to video replay) due to their high similarity.  %Since   MLP modules  dominate the latency, 
Skipping the computation-intensive MLP modules leads to substantial accelerations. 
We further provide a detailed theoretical analysis to formally characterize how our TAS controls the output differences across adjacent frames, thereby justifying the design of FastCar.

%By incorporating techniques such as KV cache compression~\cite{xiao2023streamingllm, rehg2024kvcompress, hooper2024kvquant, liu2024minicache, ge2024model, li2024snapkv} and efficient sampling methods~\cite{speculative_1, speculative_2, speculative_3, speculative_4, speculative_5, teng2024accelerating_efficient_image, he2025neighboring_efficient_ar_image, he2024zipar_efficient_image}, auto-regressive models offer a more efficient alternative to diffusion-based approaches in terms of both inference cost and memory usage, making them well-suited for high-resolution and latency-sensitive video generation.
%However, prior efficiency-oriented techniques mainly focus on attention optimization~\cite{dao2022flashattention, dao2023flashattention2, xu2025xattention_efficient_sparseattn, xi2025sparse_efficient_dit_video_sparseattn}, model compression strategies such as pruning and quantization~\cite{shen2025numrical, ma2023llmpruner, shen2024agile, lin2023awq, xiao2023smoothquant} or spatial redundancy reduction in auto-regressive image generation~\cite{he2024zipar_efficient_image, he2025neighboring_efficient_ar_image, teng2024accelerating_efficient_image}, aiming to improve inference speed and reduce memory consumption.
%In contrast, temporal redundancy—a fundamental form of redundancy inherent in video generation—remains largely overlooked, as such redundancy is less prominent in language and image generation tasks.

\textbf{Hardware Accelerator.}
We develop a flexible and efficiency-oriented hardware accelerator that supports kernel fusion and custom instruction programmability to allow direct reuse of cached outputs and enable conditional execution for MLP modules. % based on corresponding attentivity. 
%It allows direct reuse of cached outputs and eliminates redundant computation. 
To handle varying workload sparsity, we propose \textit{Dynamic Resource Scheduling (DRS)}, which leverages attentivity to dynamically allocate computational resources. % during execution. 
DRS, integrated into lightweight control logic, helps alleviate bandwidth pressure and improves overall resource efficiency, thereby enabling faster inference.
% ----------------------------------
% \asic{Meanwhile, we develop a flexible and efficiency-oriented hardware accelerator that supports kernel fusion and custom instruction programmability to enable conditional execution of MLP layers based on corresponding attentivity. This architecture allows direct reuse of cached outputs and eliminates redundant computation. To further optimize for ASICs deployment, we integrate lightweight control logic for dynamic replay, fine-grained memory access to reduce bandwidth pressure, and a configurable execution pipeline that adapts to varying workload sparsity.}
% ----------------------------------
% todo: @weize, I propose some idea for hardware optimization here: 1. direct cache usage (put cache in recent memory); 2. design lightweight control for dynamic skipping (i.e., 'replay' in this paper); 3. design configurable execution pipeline for different sparsity levels.
% dynamic resource schedule (DRS)
% dynamic computation reschdule, whole mlp now, sparse intrudce inbalance

\textbf{Comprehensive Experiments.}
Experimental results show that FastCar not only surpasses sparse attention (SA) methods with better generation quality, but also achieves faster decoding  with improved energy efficiency on FPGA.
Additionally, FastCar complements SA approaches by mitigating their drifting issues. By combining FastCar and SA, 
FastCar significantly boosts the generation quality of SA with faster inference and better long-range temporal coherence.
%Experimental results show that, for sparse attention methods, incorporating our technique not only improves generation quality and inference speed but also stabilizes long-range temporal consistency. This synergy is particularly promising for future scenarios involving higher-resolution and longer-duration video generation.}

Our contributions are summarized as follows,

\begin{itemize}[label={}, leftmargin=*]
\vspace{-2mm}
\item \textbf{1.} We perform the latency profiling and similarity analysis between different frames to explore the temporal redundancy in MLP modules. 
% Based on these specific deep insights, 
We then propose FastCar framework to accelerate AR video generation by replaying MLP modules using cached outputs from the previous frame.
% guided by the attention between tokens at corresponding spatial positions in adjacent frames.

\item \textbf{2.} Our theoretical analysis demonstrates that the similarity of MLP outputs across adjacent frames correlates with the attentivity, and this correlation is consistent across various model depths, thus justifying the design of FastCar with TAS (\textit{i.e.,} the attentivity) to guide replay decisions.

% \item \textbf{3.} \asic{We develop an efficiency-oriented hardware accelerator that enables fine-grained conditional execution of the replay, leading to faster decoding and improved energy efficiency on ASICs.}
\item \textbf{3.} We develop an efficiency-oriented hardware accelerator with DRS, enabling dynamic allocation of computational resources to enhance resource utilization and accelerate inference on FPGA.

\item \textbf{4.} Experimental results show that FastCar outperforms SA methods in generation quality and achieves more than 2.1$\times$ speedup, while also alleviating drifting issues of SA methods, thereby enhancing scalability and efficiency for high-resolution and long-duration AR video generation.

\end{itemize}

\section{Related Work}

\textbf{Auto-Regressive Visual Generation.}
Prior works~\cite{llamagen, emu3, infinity_bytedance, var_neurips, jiao2025flexvar, xie2024progressive, sun2024autoregressive, luo2024open} apply the AR framework for image generation, demonstrating its potential to outperform diffusion-based models.
In particular, VAR~\cite{var_neurips} introduces next-scale prediction to progressively generate token sequences across multiple resolutions,  demonstrating the effectiveness of  AR methods with   enhanced image quality.
Inspired by this, several works~\cite{vilau, nova, artv, kondratyuk2023videopoet} adopt the AR framework to develop  video generation models. 
However, NOVA~\cite{nova} and ART$\cdot$V~\cite{artv} still incorporate diffusion modules in their pipelines to boost generation quality, at the cost of substantially slower inference. 
Moreover, both models operate at the frame level rather than the token level, differing from the fine-grained, token-wise AR paradigm commonly used in LLMs.
Unlike the above works, VILA-U~\cite{vilau} adopts the same AR framework as LLMs,  %fully leveraging existing acceleration techniques developed for LLMs. This alignment 
making it   one of the most promising approaches in  video generation.

\textbf{Efficient Techniques for Auto-Regressive Visual Generation.}
AR image generation models~\cite{llamagen, emu3}, %trained with  next-token prediction objectives, 
typically require $n^2$ sequential forward passes to generate an image represented by $n \times n$ tokens, resulting in significant inefficiency, which is further exacerbated when extending  to video generation~\cite{vilau} with multiple image frames. 
%, where the number of tokens grows substantially with the temporal dimension.
Works~\cite{he2024zipar_efficient_image, teng2024accelerating_efficient_image} accelerate the sampling process at decode phase, utilizing contextual cues from neighboring tokens to reduce redundant computations.
The work~\cite{wang2024parallelized} trains the model from scratch to enable parallel generation of adjacent token subsets for acceleration. 
However, this approach compromises global attention modeling, which limits the generation quality.
The work~\cite{he2025neighboring_efficient_ar_image} retains a short token sequence by incorporating neighboring tokens to enable efficient generation,  thus reducing spatial redundancy.
However, these works mainly investigate spatial redundancy, leaving the temporal redundancy of video generation largely unexplored.  
%When it comes to video generation, temporal redundancy becomes significantly more critical than spatial redundancy. 
%However, none of the above works address this aspect, leaving temporal redundancy largely unexplored in this research area.

%\section{Redundancy Analysis}
\section{Deep Insights for Auto-Regressive Video Generation}

To effectively accelerate AR video generation,  we first perform detailed profiling for the inference latency and computations of VILA-U~\cite{vilau}. We then provide the following specific deep insights:  
%for AR video generation: 
(i) The decode phase takes significantly longer than the prefill  phase. 
(ii) The MLP modules dominate the latency of the decode phase. 
(iii) The outputs of an MLP module exhibit great similarity to those of its previous frame. 
Next we demonstrate our detailed observations and comprehensive analysis.

\textbf{Prefill Phase \textit{v.s.} Decode Phase.} 
We compare the latency of the prefill phase and decode phase under different input sequence lengths from 2k to 8k, during AR video generation.
As shown in the right top of Figure~\ref{fig:whole_pipeline}, the decode phase takes significantly longer than the prefill phase under various input lengths, as it needs to generate a large number of visual tokens for videos with multiple frames.
%Motivated by the bottleneck with slow decode phase,   we focus on optimizing the decode phase to accelerate auto-regressive video generation.

\textbf{Attention Modules \textit{v.s.}  MLP Modules.}  
We further explore detailed latency profiling for different decoding modules.
%There are multiple blocks with various modules (such as attention modules and MLP modules) during decoding.  
The bottom right of Figure~\ref{fig:whole_pipeline} shows that under varying input sequence lengths, MLP modules consistently dominate the overall latency. 
This observation underscores the distinct computational characteristics of AR generation compared with diffusion-based methods. 
Specifically, in diffusion transformers, all visual tokens are processed simultaneously through iterative denoising, with attention modules as the primary computational bottleneck. 
In contrast, AR models generate tokens sequentially, where attention modules only contribute marginally to the overall latency. 
As a result, efficiency-oriented techniques designed for attention modules are less effective in AR.

\begin{figure*}[t]
  \centering
  \includegraphics[width=1.0\linewidth]{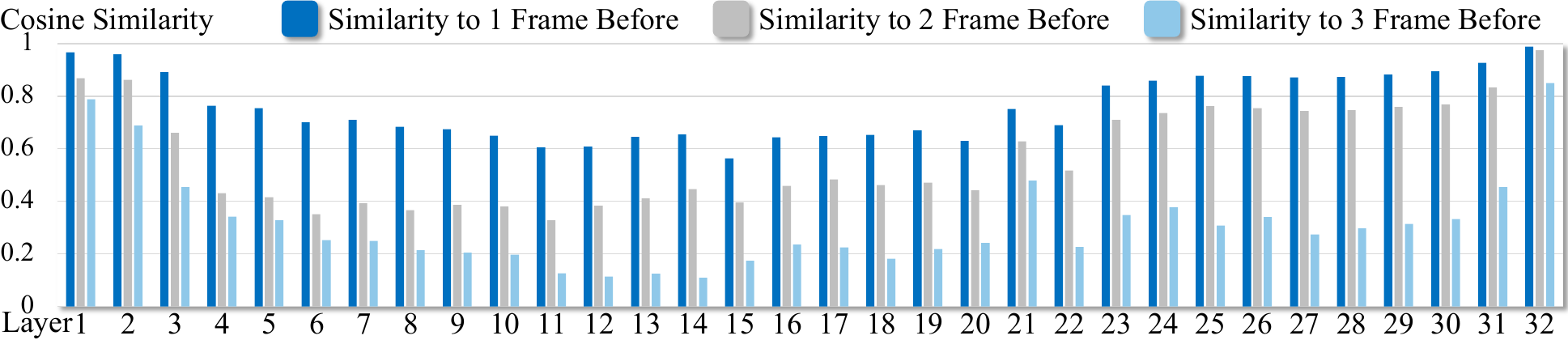}
  \caption{
  Cosine similarity for   MLP outputs between neighboring frames for all 32 MLP modules.% during the decoding phase.
  }
  \label{fig:similarity_adjacent_frames}
\end{figure*}

\textbf{Spatial Redundancy \textit{v.s.}  Temporal Redundancy.} 
Spatial Redundancy is commonly explored in image generation with SA mechanisms (to reduce computations) or efficient sampling  (to generate fewer tokens). 
%In video generation, two fundamental types of redundancy commonly arise: spatial redundancy and temporal redundancy. 
%Spatial redundancy is often explored through sparse attention mechanisms that leverage spatial patterns  in attention maps to reduce attention computations, as well as through efficient sampling techniques that utilize information from neighboring generated tokens to generate fewer tokens with less computations. 
In contrast, temporal redundancy in video generation remains largely overlooked, as prior works focus on image generation.
To explore temporal redundancy, we present cosine similarity between outputs of an MLP module and those of its neighboring frames in Figure~\ref{fig:similarity_adjacent_frames}.
The MLP outputs exhibit high similarity with their most recent frame, demonstrating high temporal redundancy.

\textbf{Motivation.} Based on the observation that MLP modules dominate the overall latency, we mainly optimize the computations of MLP modules for acceleration. The temporal redundancy with high similarities between the MLP outputs of neighboring frames further 
motivate us to cache the corresponding hidden states from the current time step for reuse in its next time step, thus avoiding its computations.

\section{FastCar Framework Design}

% \subsection{\theory{Theoretical Similarity}}
We demonstrate our FastCar framework in this section.  In general, when the proposed \textit{Temporal Attention Score (TAS)} indicates high similarity, we then directly reuse the cached outputs from its previous frame as the outputs of the current frame, thus skipping the MLP computations. 

%for each time step, we cache the MLP outputs after the attention and residual connection. Then for the next time step, if the proposed \textit{Temporal Attention Score} indicating high similarity, we can directly reuse the cached outputs from its previous time step as the outputs of this time step, thus avoiding the MLP computations.  

We first provide specific definitions for AR video generation. Then we demonstrate FastCar in great details. Theoretical analysis are further provided for the rationality and justification of   FastCar. 

\subsection{Auto-Regressive Video Generation}

% Assume the auto-regressive video generation model is denoted as function $f(\cdot)$.
% Then, the embeddings of input prompts for the model can be denoted as $X \in \mathbb{R}^{B, h, n, d}$, where $B$ denotes batch size, $h$ denotes the number of heads, $n$ denotes the sequence length of input prompt embedding, and $d$ denotes the hidden size of the model $f(\cdot)$. 

% For simplicity and better illustration, we further assume batch size $B=1$ in this section and all our results can be easily extend to other cases with a factor of $B$.

% \theory{
% TODO: 4/24 Formalize the definitions and problem and think out the form of theorem
% }

% previous_k = present_key_value[0][:, :, -257, :].unsqueeze(2)
% attn_scores = (current_q * previous_k).sum(dim=-1) / math.sqrt(128)
% B, heads=32, seq_len=1, dim=128
% We manually set the threshold \tau
% skip_mask = attn_scores > \tau
% skip some heads 
% tau is a hyperparameter

% h x d_h, hidden dim d = 128
% mlp input shape [B, 1, 4096], 4096=32x128
% self.mlp inference:  
% down_proj = self.down_proj(self.act_fn(self.gate_proj(x)) * self.up_proj(x))

% ===================================
% residual = hidden_states
% hidden_states = self.post_attention_layernorm(hidden_states)
% hidden_states = self.mlp(hidden_states)
% skip_mlp = skip_mlp.repeat(1, 128, 1).permute(0, 2, 1)
% hidden_states = torch.where(skip_mlp, self.cache_mlp[-256], hidden_states)
% hidden_states = residual + hidden_states
% ===================================      

% When computing attention, we split $X$ into $h$ heads, each with dimension $d_h$, so that $d = h \cdot d_h$.

We model a video $\mathcal{V}$ as a temporal sequence of $T$ frames with $N$ visual tokens in each frame. With $\mathcal{V}_{\text{vis}}$ denoting a finite vocabulary of visual tokens, it can be formulated as follows,
{\small
\begin{align}
\mathcal{V} = \{ z_{t,i} \mid t = 1,\dots,T;\; i = 1,\dots,N \}, \quad z_{t,i} \in \mathcal{V}_{\text{vis}},
\end{align}
}%
%where $\mathcal{V}_{\text{vis}}$ denotes a finite vocabulary of visual tokens.

Flattening the temporal–spatial grid yields a sequence of length $n = T \cdot N$.  
We denote the flattened token index as $j = (t,i) := (t{-}1)N + i$, where $t$ denotes the frame index   and $i$ denotes the index of the spatial position.
Since frames are consecutively ordered, it satisfies:
$
% \begin{align*}
(t{-}1,i) = (t,i) - N.
% \end{align*}
$

For transformer layers, we define the hidden states: $X \in \mathbb{R}^{n \times d}$, where $d$ denotes the hidden size. %Throughout this section, 
We use a batch size of $B = 1$ for simplicity, with all results readily extendable to $B > 1$ through broadcasting.
The objective of AR video generation is to model the joint distribution:
{\small
\begin{align}
P(\mathcal{V}) = \prod_{j=1}^{n} P(z_j \mid z_{<j}),
\end{align}
}%
%where $j$ denotes the index of the flattened spatiotemporal tokens.

\subsection{Key Modules}
We now formalize the key modules in AR video generation and our FastCar framework. %Similar to  a transformer-based LLM, 
The model has multiple blocks, with an attention module and an MLP module for each block, as defined below.
% \begin{definition}[Attention Module]\label{def:attn_module}
% Given hidden states $X \in \mathbb{R}^{n \times d}$, the attention projections are:
% \begin{align*}
% Q = X W_Q, \quad K = X W_K, \quad V = X W_V,
% \end{align*}
% where $W_Q, W_K, W_V \in \mathbb{R}^{d \times d}$ are the query, key, and value projection matrices.

% The attention output is computed as:
% \begin{align*}
% \mathsf{Attn}(X) = \mathsf{Softmax}\left( \frac{Q K^\top}{\sqrt{d}} \right) V \in \mathbb{R}^{n \times d}.
% \end{align*}
% \end{definition}

\begin{definition}[Attention Module]\label{def:attn_module}
Given hidden states $X \in \mathbb{R}^{n \times d}$,  attention output is computed as:
{\small
\begin{align}
\mathsf{Attn}(X) = \mathsf{Softmax}\left( \frac{Q K^\top}{\sqrt{d}} \right) V \in \mathbb{R}^{n \times d}, \text{with} \  Q = X W_Q, \ K = X W_K, \ V = X W_V,
\end{align}
}%
where $W_Q, W_K, W_V \in \mathbb{R}^{d \times d}$ are the query, key, and value projection matrices.
\end{definition}

\begin{definition}[MLP Module]\label{def:mlp_module}
Given input hidden states $X \in \mathbb{R}^{n \times d}$,
%Let $X \in \mathbb{R}^{n \times d}$ denote the input hidden states.  
the  MLP module is defined as:
{\small
\begin{align}
\mathsf{MLP}(X) = \left( \mathsf{act}(X W_G) \circ (X W_U) \right) W_D \in \mathbb{R}^{n \times d},
\end{align}
}%
where $\mathsf{act}(\cdot)$ is a non-linear activation function (e.g., SiLU), $\circ$ denotes element-wise multiplication, and $W_G, W_U \in \mathbb{R}^{d \times d_{\mathrm{ff}}}$, $W_D \in \mathbb{R}^{d_{\mathrm{ff}} \times d}$ are learned parameters with the intermediate size of $d_{\mathrm{ff}}$.
\end{definition}

Next we define TAS to measure the token similarity of adjacent frames and  guide replay decisions. 
% at the same block, and thus determine whether to skip the MLP computations based on similarity.

\begin{definition}[Temporal Attention Score]\label{def:tem_attn_score}
The \emph{temporal attention score} at spatial position $i$ and $t$-th frame is defined as the scaled dot-product between the query vector $q_j$ of token $j = (t,i)$ and the key vector $k_{j^-}$ of its aligned token $j^- = (t{-}1,i)$:
{\small
\begin{align}
s_{t,i} = \frac{ \langle q_j, k_{j^-} \rangle }{ \sqrt{d} } \in \mathbb{R}.
\end{align}
}%
%This score quantifies the temporal similarity at a fixed spatial location  to guide replay decisions.
\end{definition}

%We  need to finish the computations of the attention module and attention-based score   to determine  whether to skip the following MLP modules  in the same block. 
% The computations of the attention-based TAS are insignificant compared with MLP modules which dominate the inference latency. %, and effective for acceleration by skipping computation-intensive MLPs. 
% We show our FastCar in details next.
In our FastCar framework, due to causal decoding, TAS is obtained directly from the attention module preceding the MLP, thus there is no additional computation cost.

%We next describe how FastCar exploits temporal redundancy to skip the MLP computation.

% \paragraph{Temporal Replay via Cache.}

\subsection{Cache Attentive Replay for Fast Generation (FastCar)}

In FastCar, with TAS, we enable attentive replay in   MLP modules by manually setting a threshold $\tau$ to filter tokens of adjacent frames with higher attentivity. %between tokens $j=(t,i)$ and  $j^- = (t-1,i)$. 
When TAS is larger than $\tau$, which indicates significant temporal similarity, we skip MLP computations by reusing the outputs of its previous frame at the same spatial location.

Specifically, for each transformer block,  at frame $t-1$, for each spatial token index $i$, we cache the MLP output as follows:
{\small
\begin{align}
Y_{(t{-}1,i),:} = \mathsf{MLP} \left( \mathsf{Attn}(X) + X \right)_{(t{-}1,i),:}.
\end{align}
}%
At frame $t$, we evaluate the set of TAS $\{s_{t,i}^{(m)}\}_{m=1}^h$ between token $(t,i)$ and its aligned token $(t{-}1,i)$ across all $h$ attention heads (Definition~\ref{def:tem_attn_score}), and compute the mean score:
{\small
\begin{align}
\bar{s}_{t,i} = \frac{1}{h} \sum_{m=1}^h s_{t,i}^{(m)}.
\end{align}
}%
When the mean score exceeds a predefined threshold $\tau$, \textit{i.e.}, $\bar{s}_{t,i} \geq \tau$, 
% \begin{align*}
% \bar{s}_{t,i} \geq \tau,
% \end{align*}
we then skip the following MLP computations of this specific token $(t,i)$  and reuse the cached output for the replay in the block:
% {\small
% \begin{align*}
% Y_{(t,i),:} = Y_{(t{-}1,i),:}.
% \end{align*}
% }%
% Otherwise, we still compute:
% {\small
% \begin{align*}
% Y_{(t,i),:} = \mathsf{MLP} \left( \mathsf{Attn}(X) + X \right)_{(t,i),:}.
% \end{align*}
% }%
{\small
\begin{align}
Y_{(t,i),:} = 
\begin{cases}
Y_{(t{-}1,i),:},  &\text{if} \ \bar{s}_{t,i} \geq \tau \\
\mathsf{MLP} \left( \mathsf{Attn}(X) + X \right)_{(t,i),:},  &\text{Otherwise}
\end{cases}
.
\end{align}
}

Otherwise, we still perform the normal MLP computations. In the AR model, there are multiple  transformer blocks with the same architecture following the same computation pattern. We apply FastCar for each block.
%and  we apply the above FastCar  for each block. Since all blocks share the same architecture  following the same computation pattern,  it is easy to apply our method for all blocks and   we do not specify the exact block index above.    
This selective replay mechanism   reduces MLP computations by leveraging temporal consistency across adjacent frames.

\subsection{Theoretical Similarity Analysis}\label{sec:theoretical_similarity_analysis}

% \theory{TODO: prove the mlp output similarity for adjacent frames in the same spatical position are correlated/with proportional to attention score} 

% We now formally relate temporal attention scores to the output difference of attention outputs across adjacent frames in Theorem~\ref{thm:attn_score_dif}. 
% This result justifies that temporal attention similarity naturally enables caching with controllable performance degradation.
% \todo{TODO: this sentence is not that suitable here. Yufa: will fix when finish proof of two theorems}

We now formally characterize how TAS controls the output differences across adjacent frames, thereby justifying temporal replay based on TAS.

We first relate temporal attention similarity to the difference in attention outputs.

\begin{theorem}[Attention Score Controls Attention Output Difference]\label{thm:attn_score_dif}
Let $X \in \mathbb{R}^{n \times d}$ be the hidden states, where each row $x_j \in \mathbb{R}^d$ represents the hidden state of token $j$.
Let $\mathsf{Attn}(X)$ denote the attention output defined in Definition~\ref{def:attn_module}.  
For tokens $j = (t,i)$ and $j^- = (t{-}1,i)$ aligned at the same spatial position, define the temporal attention score $s_{t,i}$ as in Definition~\ref{def:tem_attn_score}.
Assume that:
\begin{itemize}
    \item (1) The hidden states are bounded: $\|x_j\|_2 \leq M$ for all $j$;
    \item (2) The projection matrices satisfy $\|W_Q\|_2, \|W_K\|_2 \leq \Lambda$;
    \item (3) The query and key vectors are normalized: $\|q_j\|_2 = \|k_{j^-}\|_2 = 1$ for all $j$.
\end{itemize}
% (1) The hidden states are bounded: $\|x_j\|_2 \leq M$ for all $j$;
% (2) The projection matrices satisfy $\|W_Q\|_2, \|W_K\|_2 \leq \Lambda$;
% (3) The query and key vectors are normalized: $\|q_j\|_2 = \|k_{j^-}\|_2 = 1$ for all $j$.
Let $\gamma := \|W_Q - W_K\|_2$ denote the projection difference.

Then, under the Lipschitz continuity of the attention, there exists a constant $C > 0$ such that:
{\small
\begin{align}
\| \mathsf{Attn}(X)_{j,:} - \mathsf{Attn}(X)_{j^-,:} \|_2 
\leq C\left( \sqrt{1 - s_{t,i}} + \gamma M \right).   
\end{align}
}%
Thus, a larger $s_{t,i}$ implies a smaller attention output difference up to a model-dependent offset $\gamma M$.
\end{theorem}

In practice, transformers often apply normalization techniques (such as LayerNorm or explicit vector normalization) to control query and key magnitudes. Thus assuming $\|q_j\|_2 = \|k_{j^-}\|_2 = 1$ is reasonable and standard for theoretical analysis \cite{shen2024lazydit,shen2025numrical}. 
% We provide a proof sketch here and delay the full proof to Appendix~\ref{sec:app:proofs}.
We delay the full proof to Appendix~\ref{sec:app:proofs}.

% \begin{proof}[Proof Sketch of Theorem~\ref{thm:attn_score_dif}]
% The temporal score $s_{t,i}$ measures the cosine similarity between the current query and the previous key.  
% Higher similarity implies smaller distance between query and key vectors by norm identities.  
% Using Lipschitz continuity of the attention mechanism and the linearity of key construction,  
% we bound the attention output difference up to a model-dependent offset from projection discrepancies.
% \end{proof}

\begin{remark}\label{rmk:temp_score}
The TAS $s_{t,i}$ depends only on the local query and key vectors at the current layer and is independent of model depth.  
It captures instantaneous similarity without accumulating information across layers,  
making it a stable, efficient, and fine-grained signal for dynamic computation reduction during auto-regressive decoding.
\end{remark}

Next, we relate input and attention similarity to MLP output similarity.

\begin{theorem}[Attention and Input Similarity Implies MLP Output Similarity]\label{thm:attn_mlp_dif}
Let $\mathsf{MLP}(\cdot)$ denote the MLP module defined in Definition~\ref{def:mlp_module}.  
Let $Y_{j,:} = \mathsf{MLP}\left( \mathsf{Attn}(X) + X \right)_{j,:}$ and $Y_{j^-,:} = \mathsf{MLP}\left( \mathsf{Attn}(X) + X \right)_{j^-,:}$ denote the MLP outputs at tokens $j$ and $j^-$.

Assume that $\mathsf{MLP}(\cdot)$ is $L$-Lipschitz continuous.  
Then
{\small
\begin{align}
\| Y_{j,:} - Y_{j^-,:} \|_2
\leq L \left( \| X_{j,:} - X_{j^-,:} \|_2 + \| \mathsf{Attn}(X)_{j,:} - \mathsf{Attn}(X)_{j^-,:} \|_2 \right).
\end{align}
}%
\end{theorem}

The proof is demonstrated in Appendix~\ref{sec:app:proofs}. Finally, combining the two results, we directly relate TAS to MLP output similarity.

\begin{theorem}[Temporal Attention Score Controls MLP Output Similarity]\label{thm:attn_score_mlp}
Let $X \in \mathbb{R}^{n \times d}$ be the hidden states, and let $Y_{j,:}$ and $Y_{j^-,:}$ denote the MLP outputs at tokens $j$ and $j^-$.  
Let $s_{t,i}$ denote the temporal attention score.
Under the assumptions of Theorem~\ref{thm:attn_score_dif} and assuming the MLP is $L$-Lipschitz,  
there exists a constant $C > 0$ such that:
\begin{align}
\| Y_{j,:} - Y_{j^-,:} \|_2 
\leq C\left( \|X_{j,:} - X_{j^-,:}\|_2 + \sqrt{1 - s_{t,i}} + \gamma M \right).    
\end{align}
\end{theorem}

% Together, Theorems~\ref{thm:attn_score_dif}, \ref{thm:attn_mlp_dif}, and \ref{thm:attn_score_mlp} provide a complete theoretical justification for selectively reusing MLP outputs based on temporal attention scores.
The proof is demonstrated in Appendix~\ref{sec:app:proofs}.
Theorem~\ref{thm:attn_score_mlp} formally establishes that high TAS, combined with input similarity, guarantees small MLP output deviation across adjacent frames.  
This justifies the use of thresholds on TAS to dynamically skip MLP computations during decoding, enabling efficient temporal replay with controlled quality loss.
Moreover, as TAS is computed layer-locally, it offers a stable and depth-independent signal for runtime adaptation (Remark~\ref{rmk:temp_score}).

% ======================= go through attention & mlp ===============

% down_proj = self.down_proj(self.act_fn(self.gate_proj(x)) * self.up_proj(x))

% $Y = MLP(attn(X) + X) $

% $Y' = MLP(attn(X') + X') $

% We need validate how the attention score is correlated to the similarity between $Y$ and $Y'$ at attentive location using current input $X$ not previous $X'$. 

% \theory{An interesting phenomenon is that choosing attention score $s$ as similarity is independent of depth (could try to show). Can add one line to explain this sicne $s$ does not depend on model depth.
% Add interpretation after two theorem and combining them.
% }

\section{Hardware Design}
% \subsection{Top-Level Design}

\begin{figure*}[t]
  \centering
  \includegraphics[width=1.0\linewidth]{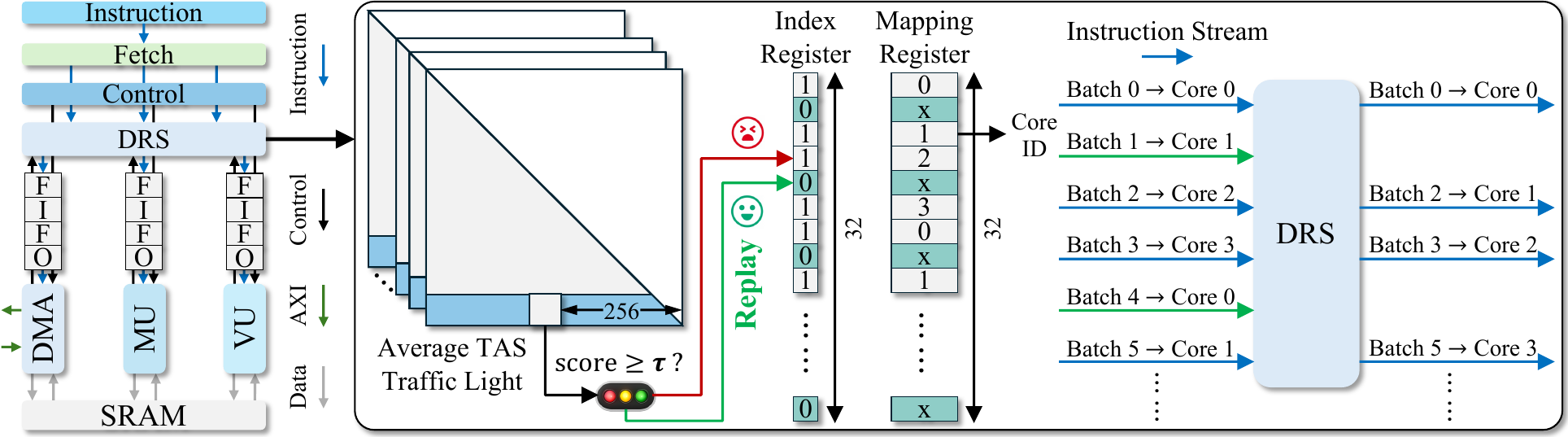}
  \caption{
  \textbf{Left}: The top-level block diagram of our hardware accelerator.
  \textbf{Right}: The DRS diagram.
  }
  \label{fig:drs}
\end{figure*}

We develop a programmable hardware accelerator, as shown in the left of Figure~\ref{fig:drs}. Pre-compiled instructions are loaded via the AXI bus with the Fetch module, and dispatched to the corresponding instruction FIFO (First-In-First-Out). % based on their status and the instruction ID.
The Control module manages the Matrix Unit (MU) and Vector Unit (VU) to perform matrix multiplication and vector computation, while the Direct Memory Access (DMA) module is responsible for loading data from off-chip memory (i.e., DDR or HBM).
The instruction FIFO receives control signals from the Control Module to coordinate the computations of each unit.
The Dynamic Resource Scheduling (DRS) module is employed to address the computational workload imbalance caused by dynamic replay from the FastCar framework.
% Our target platform is the Xilinx Alveo U280 FPGA with a chiplet design.
% We implement multiple accelerator cores on the FPGA to ensure physical implementation feasibility.

% \begin{figure*}[t]
%   \centering
%   \includegraphics[width=0.6\linewidth]{figures/top.pdf}
%   \caption{
%   The top-level block diagram of our hardware accelerator.
%   }
%   \label{fig:hardware}
% \end{figure*}

\subsection*{Dynamic Resource Scheduling (DRS)}

The FastCar framework dynamically determines whether to adopt the replay strategy to skip computations for certain MLP modules based onthe  TAS of different batches.
Due to the multi-core design, computations for different batches in dense mode are mapped to distinct cores statically.
However, the dynamic FastCar introduces computational workload imbalance across different cores, as the number of MLPS computed for different batches may vary and is difficult to predict during inference.
Moreover, we employ static compilation to pre-generate scheduling instructions. Exhaustively enumerating all possible cases would incur an unaffordable large instruction storage overhead.

% \begin{figure*}[t]
%   \centering
%   \includegraphics[width=1.0\linewidth]{figures/drs.pdf}
%   \caption{
%   The top-level block diagram of our hardware accelerator.
%   }
%   \label{fig:drs}
% \end{figure*}

To address this, we propose the DRS to balance the computational workloads, as shown in the right of Figure~\ref{fig:drs}.
After computing TAS, we configure an on-chip computation mapping table.
A 32-bit Index Register, which stores the status of each batch
(0=replay, 1=compute), is established to determine whether computation should be skipped.
32 Mapping Registers, each with $\log_2(\text{num\_cores})$ bits, determine the target core for executing which batch.
We adopt a round-robin assignment strategy to ensure balanced workload distribution among the cores.

When pre-compiled instructions are loaded and the replay mechanism is triggered, the prefetched instructions are forwarded to the DRS for processing.
The DRS then performs scheduling decisions by consulting the Index Register to determine whether to discard instructions for replayed batches or dispatch them to the appropriate cores based on the Mapping Registers. 
Notably, the DRS incurs minimal overhead by completing its dispatch operations in just hundreds to thousands of cycles, which is negligible compared to the thousands of cycles required for actual instruction execution.

\section{Experimental Results}

\subsection{Experimental Setup}
In our experiments, we mainly adopt the AR video generation model from VILA-U~\cite{vilau}  developed based on the LLaMA-2-7B~\cite{llama2} and RQ-VAE~\cite{rqvae}. The quantizer codebook size is 16384. All videos are generated with 8 frames in 256$\times$256 resolution, where each frame is decoded by   256 tokens.
VILA-U is the only available open-source model in the novel research direction of AR video generation without diffusion.
We evaluate the quality of generated videos with VBench~\cite{huang2023vbench}, and the similarity of   generated videos with metrics including  Peak Signal-to-Noise Ratio (PSNR), Structural Similarity Index Measure (SSIM), and Learned Perceptual Image Patch Similarity (LPIPS)~\cite{zhang2018perceptual}.
In detail, we compute the average similarity across all frames except the first one, as our method generates the first frame in the same manner as the baseline.
We generate videos using prompts from VBench with a batch size of 5, a fixed random seed of 42, and %a value of 3.0 for 
classifier-free guidance of 3~\cite{ho2022classifier} on A100 GPUs.
Additionally, there are no available direct baselines in this novel area, and we mainly compare our method against the sparse attention (SA) approach StreamingLLM~\cite{xiao2023streamingllm}. We set the sink size by extending the prefill length by 256 to ensure that attention to the first frame is preserved throughout video generation, thereby ensuring a fair comparison.

For hardware implementation, we adopt the Xilinx Alveo U280 FPGA as the target platform with a chiplet design.
We implement multiple accelerator cores on the FPGA to ensure physical implementation feasibility.
Latency and power are tested using a prefill sequence length of 256.

\begin{table}[]
\centering
\caption{
Main results of the proposed method compared with the sparse attention based approach StreamingLLM~\cite{xiao2023streamingllm}, where dense attention is retained in the first frame for fair comparison. Local size denotes number of local tokens for sparse attention. Detailed VBench scores are in Appendix~\ref{sec:app_additional_results}.
Latency is tested for whole generation of one video. 
Power efficiency is computed by GFLOPs/W.
}
\label{tab:main_results_compare_to_sparse_attn}
\resizebox{1.0\linewidth}{!}{
\begin{tabular}{c|cc|ccc|ccc|ccc}
\toprule
 % \multirow{2}{*}{Method} & Replay & Local & PSNR  & SSIM  & LPIPS & \multicolumn{3}{c|}{VBench Score} & \multirow{2}{*}{TMACs} & Latency & Power Effi. \\
% \multirow{2}{*}{Method} & Replay & Local & PSNR  & SSIM  & LPIPS & \multicolumn{3}{c|}{VBench Score} & TFLOPs & Latency & Power Effi. \\
\multirow{2}{*}{Method} & Replay & Local & PSNR  & SSIM  & LPIPS & \multicolumn{3}{c|}{VBench Score} & TFLOPs & Latency & Power \\
\cline{7-9}
&  Ratio  & Size  & $\uparrow$    & $\uparrow$    & $\downarrow$    & Total  & Quality & Semantic &   $\downarrow$ & (s) $\downarrow$   & Effi. $\uparrow$    \\
\midrule
Dense &  /      & /     & -     & -     & -     & 74.1\%   & 76.4\%    & 65.0\%    & 31.79
                   & 689.7 (1$\times$)   & 1.47        \\
\midrule
\multirow{5}{*}{ \shortstack{Sparse\\Attn.}} & /      & 256   & 18.25 & 51.54 & 33.59 & 72.1\%   & 74.6\%    & 62.5\%    & 30.95                   & 670.5 (1.02$\times$)   & 1.51        \\
&  /      & 128   & 13.14 & 33.61 & 54.34 & 60.7\%   & 61.9\%    & 55.9\%    & 30.82                   & 666.9 (1.03$\times$)    & 1.52        \\
& /      & 64    & 13.47 & 33.79 & 53.54 & 62.6\%   & 63.3\%    & 60.2\%    & 30.76                   & 666.3 (1.03$\times$)    & 1.52        \\
& /      & 32    & 13.34 & 33.14 & 53.82 & 61.4\%   & 61.3\%    & 62.0\%    & 30.72                   & 663.9 (1.04$\times$)   & 1.52        \\
& /      & 16    & 13.30 & 32.02 & 53.75 & 64.5\%   & 64.8\%    & 63.3\%    & 30.70                   & 662.7 (1.04$\times$)   & 1.53        \\
\midrule
\multirow{8}{*}{Ours} & 10\%   & /     & 18.57 & 53.32 & 27.31 & 73.4\%   & 75.5\%    & 65.2\%    & 30.09
                   & 629.1 (1.10$\times$)   & 1.61        \\
&  20\%   & /     & 17.94 & 51.01 & 27.57 & 73.2\%   & 75.3\%    & 65.1\%    & 28.64
                   & 556.8 (1.24$\times$)  & 1.82        \\
&  30\%   & /     & 17.87 & 50.29 & 28.02 & 72.4\%   & 74.3\%    & 64.7\%    & 27.18
                   & 525.2 (1.31$\times$)  & 1.93        \\
&  40\%   & /     & 17.68 & 50.14 & 28.15 & 71.8\%   & 73.0\%    & 67.2\%    & 25.73
                   & 487.2 (1.42$\times$)  & 2.08        \\
&  50\%   & /     & 17.85 & 50.11 & 28.08 & 71.5\%   & 72.7\%    & 66.6\%    & 24.27
                   & 475.3 (1.45$\times$)  & 2.13        \\
&  60\%   & /     & 17.85 & 50.55 & 28.72 & 71.4\%   & 72.7\%    & 66.2\%    & 22.33
                   & 451.9 (1.53$\times$)  & 2.24        \\
&  70\%   & /     & 17.86 & 50.18 & 28.79 & 71.2\%   & 72.3\%    & 66.9\%    & 20.88
                   & 415.8 (1.66$\times$)  & 2.43        \\
&  80\%   & /     & 17.71 & 49.01 & 29.50 & 71.5\%   & 73.0\%    & 65.6\%    & 19.42
                   & 390.7 (1.76$\times$)  & 2.59       \\
\bottomrule
\end{tabular}
}
\end{table}

\subsection{Main Results}

We provide the main results with different replay ratios compared with the SA method in Table~\ref{tab:main_results_compare_to_sparse_attn}. 
The detailed VBench scores for all results in different dimensions are included in Appendix~\ref{sec:app_additional_results}.
We obtain the following observations: 
(i) The SA methods, such as StreamingLLM~\cite{xiao2023streamingllm} are not able to accelerate AR video generation models effectively. When its local size shrinks with increasing sparsity, the computations measured by TFLOPs only decrease marginally without significant accelerations. The reason is that the MLP modules dominate the computations/latency, and thus optimizing attention modules does not effectively address the bottleneck. 
(ii) Our method effectively reduce the computations and achieve significant accelerations with better power efficiency. With an 80\% replay ratio, our method can reduce 45\% computations with a 1.77$\times$ acceleration.  
(iii) Our method better maintains the generation quality than the SA methods. For similarity metrics including PSNR, SSIM, and LPIPS, with gradually increasing sparsity, our generation quality only degrades marginally, which is different from the significant performance losses of SA method. For the VBench scores, we can make similar observations. Specifically, when increasing the sparsity, unlike SA method with a significant drop on the total score  (from 74.1\% to 60.7\%), our method consistently keeps a total score above 71.2\% under various sparsity levels (even with an 80\% replay ratio). 
Meanwhile, our method achieves higher power efficiency compared to the SA method, demonstrating strong potential for deployment in resource-constrained environments such as edge devices and mobile platforms.
Furthermore, we provide the visualization with our method in different replay ratios in Appendix~\ref{sec:app_additional_visualization}. 
The visualization shows that the video quality is well preserved across different replay ratios, and remains high even when the threshold $\tau$ is set to a very low value with a large replay ratio.
To summarize, FastCar maintains better generation quality with larger accelerations than SA methods.

\subsection{Ablation Study}

\begin{figure*}[t]
  \centering
  \includegraphics[width=1.0\linewidth]{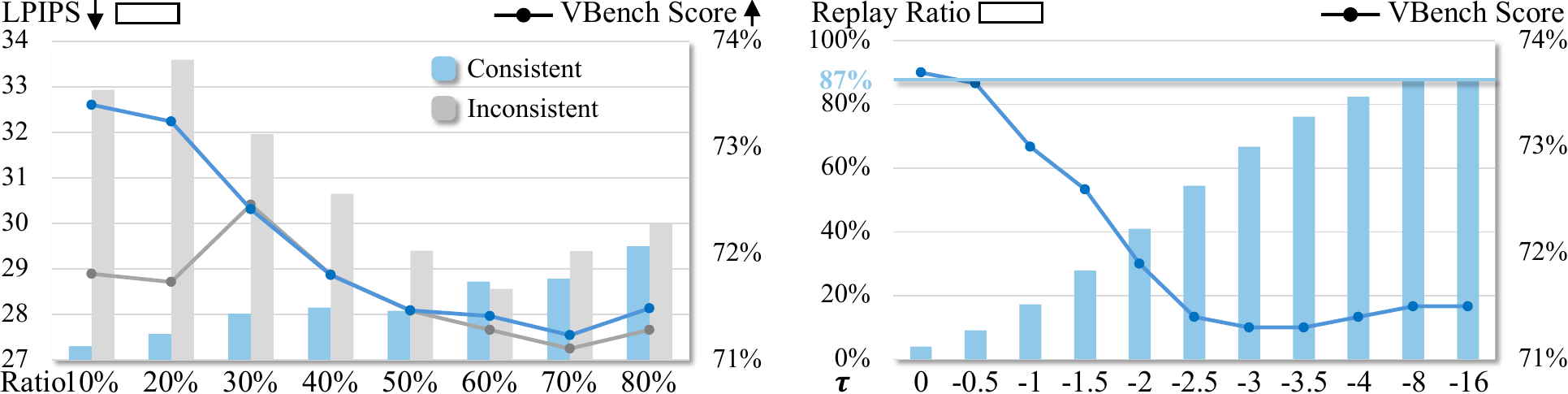}
  \caption{
  \textbf{Left:} Ablation study comparing consistent vs. inconsistent threshold settings with respect to LPIPS and the VBench total score. Full results are provided in Table~\ref{tab:supp_full_results_threshold_distribution} at Appendix~\ref{sec:app_additional_results}. 
  \textbf{Right:} Ablation study on the effect of the threshold $\tau$ on replay ratio and VBench total score. Full results are reported in Table~\ref{tab:supp_full_results_threshold_value} at Appendix~\ref{sec:app_additional_results}.
  }
  \label{fig:ablation}
  \vspace{-1mm}
\end{figure*}

\textbf{Threshold Distributions.} 
We conduct an ablation study on threshold distribution by applying either consistent or layer-wise varying (i.e., inconsistent) thresholds across all layers, while maintaining the same overall replay ratio.
The ablation results are shown in the left side of Figure~\ref{fig:ablation}.
We observe that consistent threshold achieves better performance with lower LPIPS and higher VBench score than inconsistent thresholds, which verifies the effectiveness of Remark~\ref{rmk:temp_score} discussed in Section~\ref{sec:theoretical_similarity_analysis}.

\textbf{Threshold Values.} Meanwhile, we ablate the threshold values to evaluate their impact on model performance, as shown in the right side of Figure~\ref{fig:ablation}. 
%When $\tau \le -3.5$, the replay ratio does not increase significantly. 
When $\tau \le -2.5$, if we continue to decrease $\tau$, the generation quality does not further degrade while higher sparsity with faster inference can be achieved, demonstrating the robustness of FastCar. 
Additionally, we observe that the AR video generation model achieves the highest replay ratio of 87\% when $\tau \approx -8$, indicating that only 13\% of the MLP modules are actually required during the generation process.

\begin{figure*}[t]
  \centering
  \includegraphics[width=1.0\linewidth]{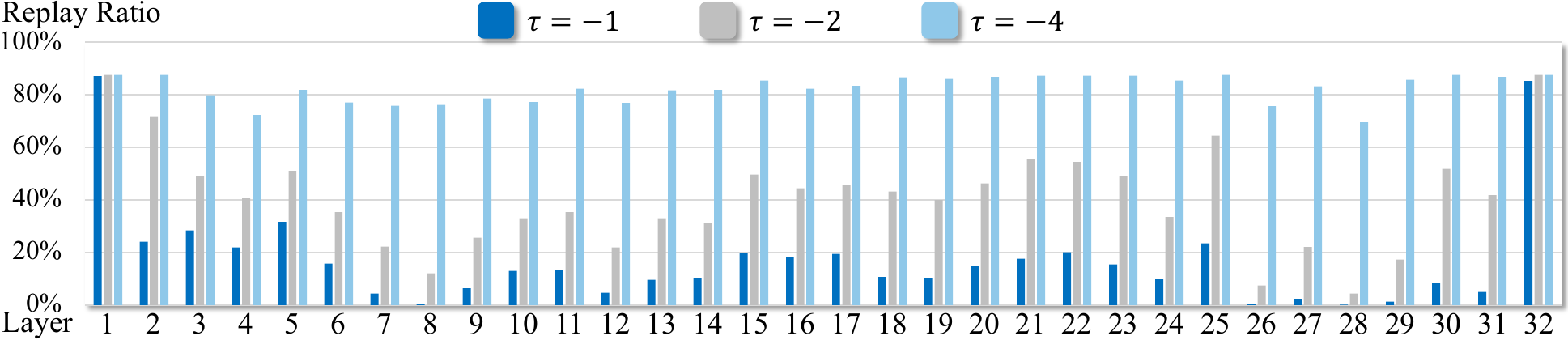}
  \caption{
  Replay ratio distribution across layers for thresholds $\tau = -1$, $-2$, and $-4$, respectively.
  }
  \label{fig:replay_ratio_distribution}
  \vspace{-1mm}
\end{figure*}

\subsection{Additional Analysis}

\paragraph{Replay Ratio Distribution.}
We visualize the replay ratio distribution across layers for 3 different thresholds in Figure~\ref{fig:replay_ratio_distribution}.
We observe that the model tends to replay at the shallow and deep layers, while replay is less likely to occur in intermediate layers.
This indicates that intermediate layers play a critical role in capturing temporal dynamics and contribute most significantly to generation quality in auto-regressive video models.

\begin{figure*}[t]
  \centering
  \includegraphics[width=1.0\linewidth]{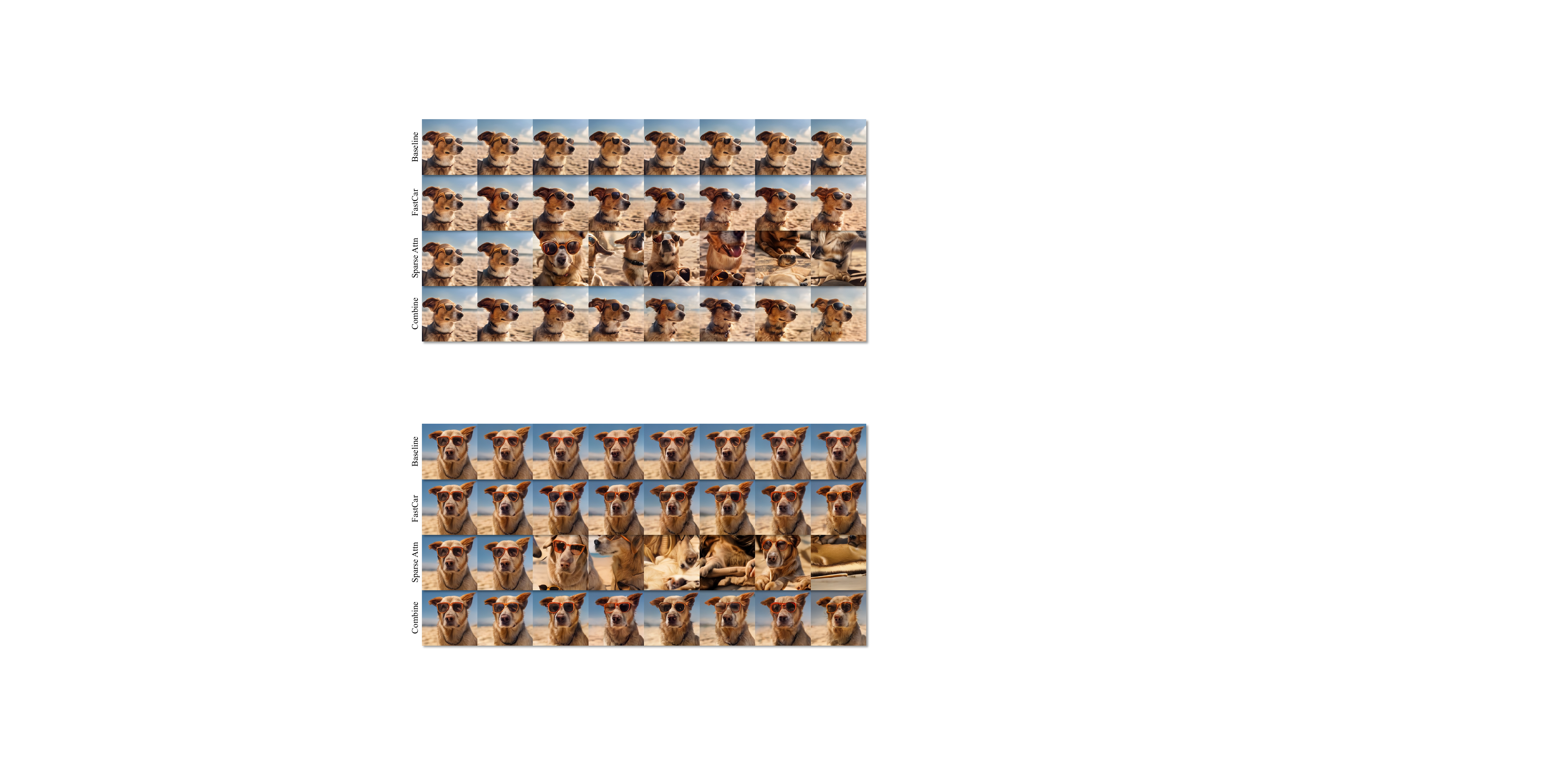}
  \caption{
  Visualization for the prompt \textit{"A dog wearing sunglasses on the beach."}. The second and third rows are generated with threshold $\tau=-4$ (i.e., 82\% replay ratio). The third and fourth rows are generated with a sink size that extends the prompt length by one frame and 64.
  }
  \vspace{-1mm}
  \label{fig:visual_ablation}
\end{figure*}

\paragraph{Combination with Sparse Attention.}
We further provide additional results achieved by combining  SA method and our method in Table~\ref{tab:additional_analysis_spattn_combine_ours}.
The detailed VBench scores for all results in different dimensions are included in Table~\ref{tab:supp_full_results_additional_analysis} at Appendix~\ref{sec:app_additional_results}.
The results show that our method can significantly boost the performance of the SA method through a straightforward combination. 
%with the support of our method, the sparse attention approach performs significantly better  than when used alone.
This validates the effectiveness of our method as a complementary enhancement to existing SA approaches.
We further visualize the results of our method, the SA approach, and their combination in Figure~\ref{fig:visual_ablation} to directly illustrate how our method alleviates drifting when integrated into sparse attention.

\begin{table}[]
\centering
\caption{
Additional results for the combination of the sparse attention method and our method. More results are included in Table~\ref{tab:supp_full_results_additional_analysis} at Appendix~\ref{sec:app_additional_results}.
}
\label{tab:additional_analysis_spattn_combine_ours}
\resizebox{1.0\linewidth}{!}{
\begin{tabular}{c|cc|ccc|ccc|ccc}
\toprule
 \multirow{2}{*}{Method} &Replay & Local & PSNR  & SSIM  & LPIPS & \multicolumn{3}{c|}{VBench Score} & GFLOPs & Latency & Power \\
\cline{7-9}
&  Ratio  & Size  & $\uparrow$    & $\uparrow$    & $\downarrow$    & Total  & Quality & Semantic &    $\downarrow$  &(s)$\downarrow$     & Effi.$\uparrow$    \\
\midrule
Dense &  /      & /     & -     & -     & -     & 74.1\%   & 76.4\%    & 65.0\%    & 31.79
                   & 689.7 (1$\times$)   & 1.47        \\
\midrule
% 87\%   & 16    & 17.27 & 46.49 & 32.37 & 71.6\%   & 73.1\%    & 65.7\%    & 29.51                   & 608.46    & 3.26        \\
% 87\%   & 32    & 17.27 & 46.75 & 31.96 & 71.9\%   & 73.4\%    & 65.9\%    & 29.54                   & 618.45    & 3.21        \\
% 87\%   & 64    & 17.27 & 46.70 & 32.09 & 71.6\%   & 73.1\%    & 65.5\%    & 29.60                   & 625.59    & 3.18        \\
% 87\%   & 128   & 17.29 & 46.79 & 32.10 & 71.7\%   & 73.3\%    & 65.7\%    & 29.73                   & 643.31    & 3.09        \\
% 87\%   & 256   & 17.44 & 47.57 & 31.27 & 71.8\%   & 73.3\%    & 65.7\%    & 29.99                   & 676.85    & 2.93        \\
\multirow{5}{*}{\shortstack{Ours\\+\\Sparse\\Attn.}} & 87\%   & 256   & 17.44 & 47.57 & 31.27 & 71.8\%   & 73.3\%    & 65.7\%    & 17.61                   & 354.46 (1.95$\times$)    & 2.85        \\
&  87\%   & 128   & 17.29 & 46.79 & 32.10 & 71.7\%   & 73.3\%    & 65.7\%    & 17.49                   & 342.25 (2.02$\times$)    & 2.96        \\
&  87\%   & 64    & 17.27 & 46.70 & 32.09 & 71.6\%   & 73.1\%    & 65.5\%    & 17.43                   & 334.57 (2.06$\times$)   & 3.02        \\
&  87\%   & 32    & 17.27 & 46.75 & 31.96 & 71.9\%   & 73.4\%    & 65.9\%    & 17.40                   & 327.36 (2.11$\times$)   & 3.09        \\
&  87\%   & 16    & 17.27 & 46.49 & 32.37 & 71.6\%   & 73.1\%    & 65.7\%    & 17.39                   & 324.31 (2.13$\times$)   & 3.12        \\
\bottomrule
\end{tabular}
}
\end{table}

\section{Conclusion}

We propose FastCar, a framework to accelerate AR video generation on the edge. 
We show that the similarity across adjacent frames correlates with attentivity and is independent with model depths. 
We then introduce a replay strategy that reuses cached MLP outputs from the previous frame to reduce computation. 
Also, the DRS design is adopted to improve resource utilization and inference speed on FPGAs.
Results show that our method outperforms prior SA approaches and complements them with more than 2.1$\times$ speedup, enabling better scalability for high-resolution, long-duration video generation.
In future work, we plan to extend our framework to a broader range of model families.

\bibliographystyle{unsrt}
\bibliography{reference}

\clearpage
\newpage

\section*{Appendix}

\section{Additional Results}\label{sec:app_additional_results}

\subsection{Detailed Results for VBench}
We provide the detailed scores of VBench in Table~\ref{tab:supp_detailed_vbench_scores_1} and Table~\ref{tab:supp_detailed_vbench_scores_2}.
Our method better maintains the generation quality than SA methods.  Specifically, on the  VBench,  when increasing the sparsity, unlike SA method with a significant drop on most of subtasks, our method keeps high scores close to the dense model on most subtasks  under various sparsity.

\begin{table}[h]
\centering
\caption{
Detailed VBench scores.
}
\label{tab:supp_detailed_vbench_scores_1}
\resizebox{1.0\linewidth}{!}{
\begin{tabular}{c|c|c|cccccccc}
\toprule
Method & Replay & Local & Overall     & Subject     & Background  & Temporal   & Motion     & Dynamic & Aesthetic & Imaging \\
& Ratio  & Size  & Consistency & Consistency & Consistency & Flickering & Smoothness & Degree  & Quality   & Quality \\
\midrule
Dense &  /      & /     & 27.9\%      & 87.8\%      & 94.6\%      & 95.8\%     & 94.9\%     & 59.4\%  & 57.4\%    & 58.8\%  \\ \midrule
\multirow{5}{*}{ \shortstack{Sparse\\Attn.}} & /      & 16    & 28.0\%      & 71.9\%      & 88.0\%      & 84.5\%     & 84.8\%     & 100.0\% & 54.2\%    & 59.1\%  \\
& /      & 32    & 27.9\%      & 66.8\%      & 86.2\%      & 82.5\%     & 82.3\%     & 100.0\% & 54.2\%    & 59.0\%  \\
& /      & 64    & 27.7\%      & 67.4\%      & 86.3\%      & 84.6\%     & 84.4\%     & 100.0\% & 54.0\%    & 58.0\%  \\
& /      & 128   & 27.5\%      & 63.7\%      & 85.1\%      & 84.4\%     & 84.3\%     & 99.7\%  & 52.9\%    & 57.2\%  \\
& /      & 256   & 27.8\%      & 82.7\%      & 92.8\%      & 92.5\%     & 92.4\%     & 94.4\%  & 55.9\%    & 56.8\%  \\
\midrule
\multirow{5}{*}{Ours} & 10\%   & /     & 27.7\%      & 86.2\%      & 94.0\%      & 95.9\%     & 94.6\%     & 57.8\%  & 57.0\%    & 57.5\%  \\
& 20\%   & /     & 27.9\%      & 86.7\%      & 93.7\%      & 95.5\%     & 94.7\%     & 57.5\%  & 56.9\%    & 57.4\%  \\
& 30\%   & /     & 27.9\%      & 85.7\%      & 93.4\%      & 95.1\%     & 94.4\%     & 52.8\%  & 57.0\%    & 57.1\%  \\
& 40\%   & /     & 27.9\%      & 87.1\%      & 93.7\%      & 95.0\%     & 94.9\%     & 35.6\%  & 57.2\%    & 57.5\%  \\
& 50\%   & /     & 27.9\%      & 88.2\%      & 94.1\%      & 95.1\%     & 95.0\%     & 20.6\%  & 57.7\%    & 57.8\%  \\
& 60\%   & /     & 27.9\%      & 89.3\%      & 94.2\%      & 94.9\%     & 95.2\%     & 12.8\%  & 57.9\%    & 58.0\%  \\
& 70\%   & /     & 28.1\%      & 89.4\%      & 94.1\%      & 94.4\%     & 95.0\%     & 11.4\%  & 58.2\%    & 58.3\%  \\
& 80\%   & /     & 28.2\%      & 88.7\%      & 93.8\%      & 93.7\%     & 94.4\%     & 23.3\%  & 58.2\%    & 58.5\%  \\
\bottomrule
\end{tabular}
}
\end{table}
\begin{table}[h]
\centering
\caption{
Detailed VBench scores.
}
\label{tab:supp_detailed_vbench_scores_2}
\resizebox{0.9\linewidth}{!}{
\begin{tabular}{c|c|c|cccccccc}
\toprule
Method & Replay & Local & Object & Multiple & Human  & \multirow{2}{*}{Color} & Spatial      & \multirow{2}{*}{Scene} & Appearance & Temporal \\
& Ratio  & Size  & Class  & Objects  & Action &                        & Relationship &                        & Style      & Style    \\
\midrule
Dense& /      & /     & 76.7\% & 30.8\%   & 74.8\% & 82.0\% & 37.6\%       & 42.6\% & 24.7\%     & 25.0\%   \\  \midrule
\multirow{5}{*}{ \shortstack{Sparse\\Attn.}} 
& /      & 16    & 67.8\% & 20.5\%   & 81.8\% & 81.7\% & 41.2\%       & 37.3\% & 24.6\%     & 24.9\%   \\
& /      & 32    & 65.7\% & 19.0\%   & 83.2\% & 80.0\% & 37.2\%       & 35.0\% & 24.5\%     & 24.8\%   \\
& /      & 64    & 61.9\% & 17.2\%   & 81.8\% & 77.6\% & 30.9\%       & 33.7\% & 24.5\%     & 25.2\%   \\
& /      & 128   & 50.1\% & 12.0\%   & 78.8\% & 73.4\% & 19.9\%       & 32.6\% & 24.3\%     & 25.1\%   \\
& /      & 256   & 68.1\% & 26.6\%   & 80.4\% & 76.7\% & 31.8\%       & 38.9\% & 24.5\%     & 25.3\%   \\
\midrule
\multirow{5}{*}{Ours} & 10\%   & /     & 74.9\% & 33.7\%   & 76.8\% & 78.6\% & 42.2\%       & 40.4\% & 24.7\%     & 24.9\%   \\
& 20\%   & /     & 74.0\% & 34.0\%   & 76.0\% & 78.9\% & 41.0\%       & 40.6\% & 24.7\%     & 25.1\%   \\
& 30\%   & /     & 74.4\% & 32.8\%   & 75.8\% & 78.0\% & 39.8\%       & 40.8\% & 24.8\%     & 25.1\%   \\
& 40\%   & /     & 76.5\% & 31.0\%   & 73.8\% & 76.1\% & 40.9\%       & 40.6\% & 24.7\%     & 25.2\%   \\
& 50\%   & /     & 77.1\% & 34.3\%   & 74.0\% & 76.8\% & 42.7\%       & 42.0\% & 24.8\%     & 25.2\%   \\
& 60\%   & /     & 77.6\% & 36.0\%   & 75.4\% & 79.2\% & 43.5\%       & 42.2\% & 24.8\%     & 25.3\%   \\
& 70\%   & /     & 77.9\% & 36.3\%   & 79.0\% & 79.2\% & 43.8\%       & 42.6\% & 24.8\%     & 25.3\%   \\
& 80\%   & /     & 77.9\% & 36.3\%   & 75.4\% & 77.3\% & 44.9\%       & 42.3\% & 24.7\%     & 25.4\%  \\
\bottomrule
\end{tabular}
}
\end{table}

\subsection{Detailed Ablation Results for Threshold Distribution}
We provide full results for the ablation of threshold distribution in Table~\ref{tab:supp_full_results_threshold_distribution}.  We observe that  consistent threshold  achieves better performance with lower LPIPS and higher VBench score than  inconsistent thresholds, which verifies the effectiveness of Remark~\ref{rmk:temp_score} discussed in Section~\ref{sec:theoretical_similarity_analysis}.

\subsection{Detailed Ablation Results for Threshold Values}
We provide full results for the ablation of threshold values in Table~\ref{tab:supp_full_results_threshold_value}. 
When $\tau \le -2.5$, if we continue to decrease $\tau$, the generation quality does not further degrade while higher sparsity with faster inference is achieved, demonstrating the robustness of FastCar. 
Additionally, we observe that the AR video generation model achieves the highest replay ratio of 87\% when $\tau \approx -8$, indicating that only 13\% of the MLP modules are actually required during the generation process.

\subsection{Full Results for Additional Analysis}
We provide full results for the combination of the sparse attention method and our method in Table~\ref{tab:supp_full_results_additional_analysis}. The results show that our method significantly boosts the performance of SA method through the straightforward combination. 
This validates the effectiveness of our method as a complementary enhancement to existing SA approaches.

\begin{table}[h]
\centering
\caption{
Full results for the ablation of the threshold distribution.
}
\label{tab:supp_full_results_threshold_distribution}
\resizebox{0.7\linewidth}{!}{
\begin{tabular}{c|c|ccc|ccc}
\toprule
Threshold   & Replay & PSNR  & SSIM  & LPIPS & \multicolumn{3}{c}{VBench Score} \\
\cline{6-8}
Distribution    & Ratio  & $\uparrow$    & $\uparrow$    & $\downarrow$    & Total    & Quality   & Semantic  \\
\midrule
Consistent     & 10\%   & 18.57 & 53.32 & 27.31 & 73.4\%   & 75.5\%    & 65.2\%    \\
Inconsistent & 10\%   & 16.73 & 46.63 & 32.94 & 71.8\%   & 73.7\%    & 64.3\%    \\
\midrule
Consistent     & 20\%   & 17.94 & 51.01 & 27.57 & 73.2\%   & 75.3\%    & 65.1\%    \\
Inconsistent & 20\%   & 16.30 & 45.05 & 33.60 & 71.7\%   & 73.3\%    & 65.4\%    \\
\midrule
Consistent     & 30\%   & 17.87 & 50.29 & 28.02 & 72.4\%   & 74.3\%    & 64.7\%    \\
Inconsistent & 30\%   & 16.67 & 45.39 & 31.96 & 72.5\%   & 74.0\%    & 66.5\%    \\
\midrule
Consistent     & 40\%   & 17.68 & 50.14 & 28.15 & 71.8\%   & 73.0\%    & 67.2\%    \\
Inconsistent & 40\%   & 17.34 & 48.61 & 30.65 & 71.8\%   & 73.6\%    & 64.5\%    \\
\midrule
Consistent     & 50\%   & 17.85 & 50.11 & 28.08 & 71.5\%   & 72.7\%    & 66.6\%    \\
Inconsistent & 50\%   & 17.65 & 49.94 & 29.40 & 71.5\%   & 73.0\%    & 65.4\%    \\
\midrule
Consistent     & 60\%   & 17.85 & 50.55 & 28.72 & 71.4\%   & 72.7\%    & 66.2\%    \\
Inconsistent & 60\%   & 17.79 & 49.55 & 28.56 & 71.3\%   & 72.5\%    & 66.5\%    \\
\midrule
Consistent     & 70\%   & 17.86 & 50.18 & 28.79 & 71.2\%   & 72.3\%    & 66.9\%    \\
Inconsistent & 70\%   & 17.68 & 48.84 & 29.39 & 71.1\%   & 72.4\%    & 65.9\%    \\
\midrule
Consistent     & 80\%   & 17.71 & 49.01 & 29.50 & 71.5\%   & 73.0\%    & 65.6\%    \\
Inconsistent & 80\%   & 17.59 & 48.06 & 30.01 & 71.3\%   & 72.5\%    & 66.3\%   \\
\bottomrule
\end{tabular}
}
\end{table}
\begin{table}[h]
\centering
\caption{
Full results for the ablation of the threshold values.
}
\label{tab:supp_full_results_threshold_value}
\resizebox{0.7\linewidth}{!}{
\begin{tabular}{c|c|ccc|ccc}
\toprule
Threshold & Replay  & PSNR  & SSIM  & LPIPS & \multicolumn{3}{c}{VBench Score} \\
\cline{6-8}
Values    & Ratio   & $\uparrow$    & $\uparrow$    & $\downarrow$    & Total    & Quality   & Semantic  \\
\midrule
0         & 3.96\%  & 19.71 & 57.66 & 24.14 & 73.7\%   & 75.7\%    & 65.8\%    \\
-0.5      & 9.13\%  & 18.61 & 53.49 & 27.27 & 73.6\%   & 75.6\%    & 65.5\%    \\
-1        & 17.32\% & 17.84 & 50.49 & 29.22 & 73.0\%   & 75.1\%    & 64.6\%    \\
-1.5      & 27.81\% & 17.31 & 48.39 & 30.85 & 72.6\%   & 74.5\%    & 64.7\%    \\
-2        & 40.92\% & 17.38 & 48.84 & 30.45 & 71.9\%   & 73.6\%    & 65.2\%    \\
-2.5      & 54.55\% & 17.76 & 50.30 & 29.05 & 71.4\%   & 72.7\%    & 65.8\%    \\
-3        & 66.78\% & 17.87 & 50.42 & 28.69 & 71.3\%   & 72.5\%    & 66.3\%    \\
-3.5      & 76.20\% & 17.76 & 49.42 & 29.26 & 71.3\%   & 72.6\%    & 66.2\%    \\
-4        & 82.41\% & 17.65 & 48.51 & 29.83 & 71.4\%   & 72.7\%    & 66.2\%    \\
-8        & 87.49\% & 17.60 & 48.50 & 30.09 & 71.5\%   & 72.9\%    & 66.0\%    \\
-16       & 87.49\% & 17.60 & 48.04 & 30.09 & 71.5\%   & 72.9\%    & 66.0\%   \\
\bottomrule
\end{tabular}
}
\end{table}

\begin{table}[h]
\centering
\caption{
Full results for the combination of the sparse attention method and our method.
}
\label{tab:supp_full_results_additional_analysis}
\resizebox{1.0\linewidth}{!}{
\begin{tabular}{c|c|c|c|ccc|ccc|ccc}
\toprule
% Method & Threshold & Replay  & Local & PSNR  & SSIM  & LPIPS & \multicolumn{3}{c|}{VBench Score} & \multirow{2}{*}{TMACs} & Latency & Power Effi. \\
Method & Threshold & Replay  & Local & PSNR  & SSIM  & LPIPS & \multicolumn{3}{c|}{VBench Score} & TFLOPs & Latency & Power \\
\cline{8-10}
% & Value     & Ratio   & Size  & $\uparrow$    & $\uparrow$    & $\downarrow$    & Total    & Quality   & Semantic  &                        & (s)     & GMACs/W     \\
& Value     & Ratio   & Size  & $\uparrow$    & $\uparrow$    & $\downarrow$    & Total    & Quality   & Semantic  &                    $\downarrow$    & (s) $\downarrow$     & Effi. $\uparrow$     \\
\midrule
Dense & /  & /       & /   & -     & -     & -     & 74.1\% & 76.4\% & 65.0\% & 31.79 & 689.71 & 1.47 \\
\midrule
\multirow{2}{*}{\shortstack{Ours\\+\\Sparse\\Attn.}} & -1 & 17.72\% & 16  & 12.96 & 29.57 & 55.13 & 60.8\% & 61.6\% & 57.8\% & 28.05 & 497.35 & 2.03 \\
& -2 & 46.11\% & 16  & 14.38 & 36.14 & 47.50 & 64.7\% & 66.5\% & 57.5\% & 23.69 & 427.75 & 2.36 \\
&-3 & 70.23\% & 16  & 16.95 & 45.72 & 34.25 & 70.5\% & 72.0\% & 64.9\% & 19.81 & 356.35 & 2.84 \\
&-4 & 83.85\% & 16  & 17.27 & 46.49 & 32.37 & 71.6\% & 73.1\% & 65.7\% & 17.87 & 331.29 & 3.05 \\
\midrule
\multirow{2}{*}{\shortstack{Ours\\+\\Sparse\\Attn.}} & -1 & 17.72\% & 32  & 13.25 & 31.58 & 53.77 & 61.2\% & 61.9\% & 58.2\% & 28.07 & 499.79 & 2.02 \\
& -2 & 46.11\% & 32  & 14.43 & 36.81 & 47.18 & 65.0\% & 66.6\% & 59.0\% & 23.71 & 430.18 & 2.35 \\
& -3 & 70.23\% & 32  & 16.94 & 46.02 & 34.09 & 70.8\% & 72.3\% & 64.9\% & 19.83 & 358.78 & 2.82 \\
& -4 & 83.85\% & 32  & 17.27 & 46.75 & 31.96 & 71.9\% & 73.4\% & 65.9\% & 17.89 & 333.73 & 3.03 \\
\midrule
\multirow{2}{*}{\shortstack{Ours\\+\\Sparse\\Attn.}} & -1 & 17.72\% & 64  & 13.25 & 31.90 & 54.32 & 60.0\% & 60.9\% & 56.5\% & 28.10 & 504.60 & 2.00 \\
& -2 & 46.11\% & 64  & 14.41 & 36.83 & 47.72 & 64.6\% & 66.2\% & 58.4\% & 23.74 & 435.00 & 2.33 \\
& -3 & 70.23\% & 64  & 16.88 & 45.87 & 34.60 & 70.4\% & 71.8\% & 64.7\% & 19.86 & 363.60 & 2.78 \\
& -4 & 83.85\% & 64  & 17.27 & 46.70 & 32.09 & 71.6\% & 73.1\% & 65.5\% & 17.92 & 338.55 & 2.99 \\
\midrule
\multirow{2}{*}{\shortstack{Ours\\+\\Sparse\\Attn.}} &  -1 & 17.72\% & 128 & 13.14 & 31.63 & 55.02 & 59.1\% & 60.3\% & 53.9\% & 28.16 & 514.00 & 1.97 \\
& -2 & 46.11\% & 128 & 14.44 & 37.20 & 47.80 & 64.2\% & 65.9\% & 57.6\% & 23.80 & 444.40 & 2.28 \\
& -3 & 70.23\% & 128 & 16.89 & 46.02 & 34.77 & 70.0\% & 71.5\% & 63.9\% & 19.92 & 373.00 & 2.71 \\
& -4 & 83.85\% & 128 & 17.29 & 46.79 & 32.10 & 71.7\% & 73.3\% & 65.7\% & 17.98 & 347.95 & 2.91 \\
\midrule
\multirow{2}{*}{\shortstack{Ours\\+\\Sparse\\Attn.}} &  -1 & 17.72\% & 256 & 15.22 & 40.77 & 44.61 & 67.7\% & 70.0\% & 58.7\% & 28.28 & 531.87 & 1.90 \\
& -2 & 46.11\% & 256 & 15.78 & 43.05 & 40.25 & 68.9\% & 71.1\% & 60.1\% & 23.92 & 462.27 & 2.19 \\
& -3 & 70.23\% & 256 & 17.21 & 47.63 & 32.94 & 70.7\% & 72.2\% & 64.8\% & 20.04 & 390.87 & 2.59 \\
& -4 & 83.85\% & 256 & 17.44 & 47.57 & 31.27 & 71.8\% & 73.3\% & 65.7\% & 18.10 & 365.82 & 2.76 \\
\bottomrule
\end{tabular}
}
\end{table}

\clearpage
\newpage
\section{Additional Visualization}\label{sec:app_additional_visualization}

We  visualize the results of our method under different replay ratios. Our method generates high quality videos. 

\begin{figure*}[h]
  \centering
  \includegraphics[width=0.9\linewidth]{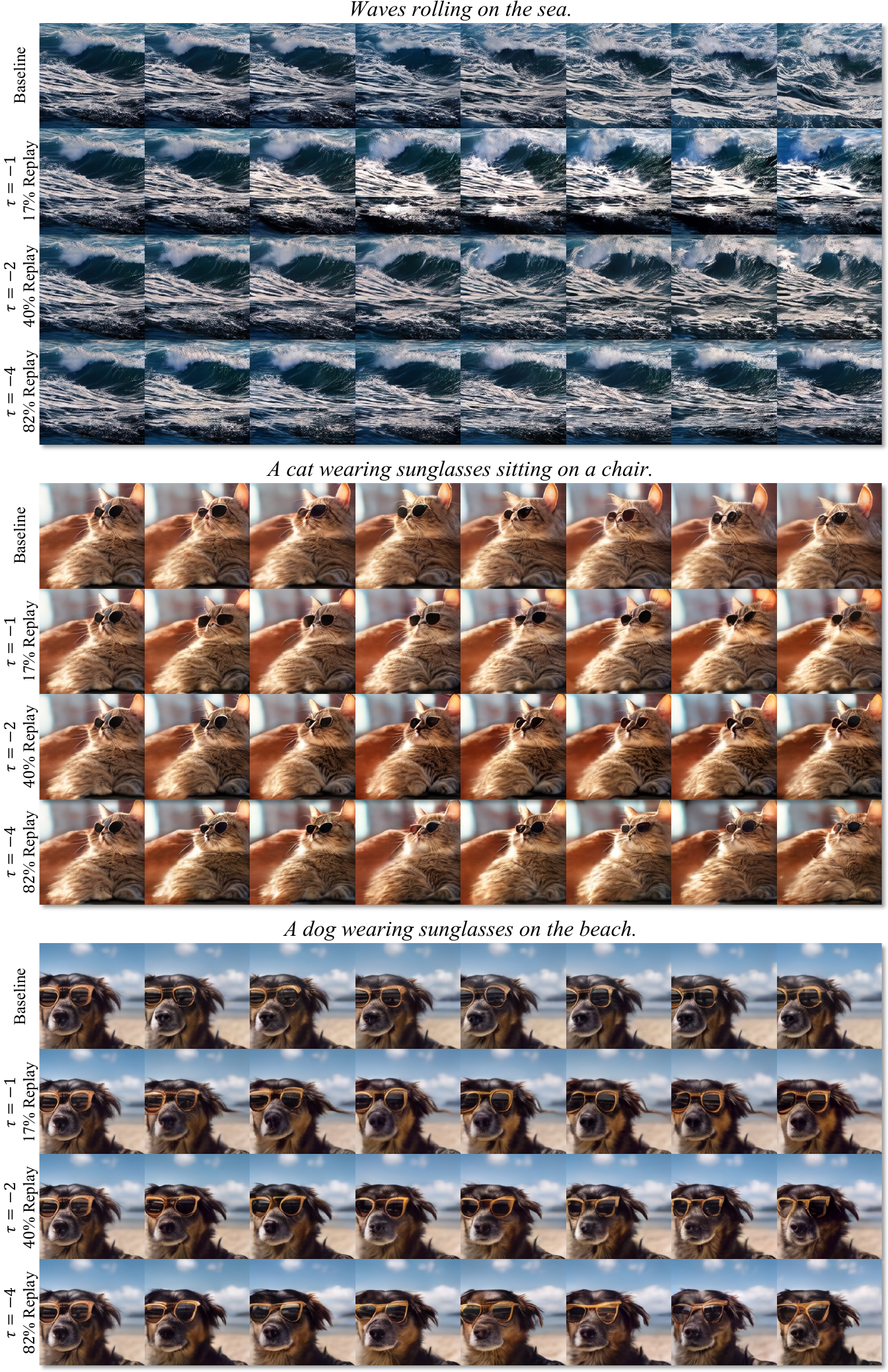}
  \caption{
  Additional visualization with threshold $\tau=-1,-2,-4$.
  }
  \label{fig:supp_visualization_different_replay_ratios}
\end{figure*}

% \begin{figure*}[t]
%   \centering
%   \includegraphics[width=1.0\linewidth]{figures/visualization-ablation-supp.pdf}
%   \caption{
%   Additional visualization for the prompt \textit{"A dog wearing sunglasses on the beach."}. The results at second and third row is generated with threshold $\tau=-4$ (i.e., 82\% replay ratio); the results at third and fourth rows are generated with with 64 sink size.
%   }
%   \label{fig:visual_ablation}
% \end{figure*}

\clearpage
\newpage
\section{Detailed Proofs}\label{sec:app:proofs}

\subsection{Proof of Theorem~\ref{thm:attn_score_dif}}
\begin{proof}[Proof of Theorem~\ref{thm:attn_score_dif}]
\textbf{Step 1 (Score exactly matches cosine similarity).}
By Definition~\ref{def:tem_attn_score}, $s_{t,i} = \langle q_j, k_{j^-} \rangle / \sqrt{d}$, where $q_j = x_j W_Q$ and $k_{j^-} = x_{j^-} W_K$.  
Under Assumption (3), $\|q_j\|_2 = \|k_{j^-}\|_2 = 1$, so $s_{t,i}$ (up to $\sqrt{d}$ scaling) equals the cosine similarity:
\[
\cos\theta(q_j, k_{j^-}) = \langle q_j, k_{j^-} \rangle.
\]
Thus, by the Law of Cosines for unit vectors,
\[
\| q_j - k_{j^-} \|_2^2 = 2(1 - s_{t,i}).
\]

\textbf{Step 2 (Logit gap from query gap).}
The attention logits satisfy
\[
\ell_j = q_j K^\top, \quad \ell_{j^-} = q_{j^-} K^\top,
\]
thus
\begin{align*}
\| \ell_j - \ell_{j^-} \|_2 
&= \| (q_j - q_{j^-}) K^\top \|_2 \\
&\leq \|K\|_2 \| q_j - q_{j^-} \|_2,
\end{align*}
where $K = X W_K$ is the stacked key matrix.  
Since $K = X W_K$, we have
\[
\|K\|_2 \leq \|X\|_2 \|W_K\|_2 \leq \sqrt{n} M \Lambda,
\]
where $\|X\|_2 \leq \sqrt{n} M$ since each $\|x_j\|_2 \leq M$.

\textbf{Step 3 (Attention output is Lipschitz).}
Since softmax and value projection are Lipschitz continuous (see \cite{shen2024lazydit}), there exists $L_{\text{attn}} > 0$ such that
\[
\| \mathsf{Attn}(X)_{j,:} - \mathsf{Attn}(X)_{j^-,:} \|_2 
\leq L_{\text{attn}} \| \ell_j - \ell_{j^-} \|_2
\leq C_1 \| q_j - q_{j^-} \|_2,
\]
where $C_1 = L_{\text{attn}} \sqrt{n} M \Lambda$.

\textbf{Step 4 (Bounding query–key difference).}
Since
\[
q_{j^-} = x_{j^-} W_Q, \quad k_{j^-} = x_{j^-} W_K,
\]
it follows that
\[
\| k_{j^-} - q_{j^-} \|_2 
= \| x_{j^-} (W_K - W_Q) \|_2 
\leq \gamma \| x_{j^-} \|_2 
\leq \gamma M.
\]
By triangle inequality,
\[
\| q_j - q_{j^-} \|_2 
\leq \| q_j - k_{j^-} \|_2 + \| k_{j^-} - q_{j^-} \|_2
\leq \sqrt{2(1-s_{t,i})} + \gamma M.
\]

\textbf{Step 5 (Final bound).}
Thus,
\begin{align*}
\| \mathsf{Attn}(X)_{j,:} - \mathsf{Attn}(X)_{j^-,:} \|_2 
&\leq C_1 \left( \sqrt{2(1-s_{t,i})} + \gamma M \right) \\
&\leq C\left( \sqrt{1-s_{t,i}} + \gamma M \right),
\end{align*}
after absorbing constants into $C>0$.
This completes the proof.
\end{proof}

\subsection{Proof of Theorem~\ref{thm:attn_mlp_dif}}
\begin{proof}[Proof of Theorem~\ref{thm:attn_mlp_dif}]
Define
\[
Z_j = \mathsf{Attn}(X)_{j,:} + X_{j,:},
\quad
Z_{j^-} = \mathsf{Attn}(X)_{j^-,:} + X_{j^-,:}.
\]
Then
\[
Y_{j,:} = \mathsf{MLP}(Z_j), \quad Y_{j^-,:} = \mathsf{MLP}(Z_{j^-}).
\]
By Lipschitz continuity of $\mathsf{MLP}$,
\[
\| Y_{j,:} - Y_{j^-,:} \|_2 \leq L \| Z_j - Z_{j^-} \|_2.
\]
Expanding $Z_j - Z_{j^-}$ and applying triangle inequality,
\[
\| Z_j - Z_{j^-} \|_2 \leq \| \mathsf{Attn}(X)_{j,:} - \mathsf{Attn}(X)_{j^-,:} \|_2 + \| X_{j,:} - X_{j^-,:} \|_2.
\]
The claim follows.
\end{proof}

\subsection{Proof of Theorem~\ref{thm:attn_score_mlp}}
\begin{proof}[Proof of Theorem~\ref{thm:attn_score_mlp}]
By Theorem~\ref{thm:attn_mlp_dif},
\[
\| Y_{j,:} - Y_{j^-,:} \|_2
\leq L\left( \| X_{j,:} - X_{j^-,:} \|_2 + \| \mathsf{Attn}(X)_{j,:} - \mathsf{Attn}(X)_{j^-,:} \|_2 \right).
\]
By Theorem~\ref{thm:attn_score_dif},
\[
\| \mathsf{Attn}(X)_{j,:} - \mathsf{Attn}(X)_{j^-,:} \|_2
\leq C'\left( \sqrt{1 - s_{t,i}} + \gamma M \right).
\]
Substituting gives
\[
\| Y_{j,:} - Y_{j^-,:} \|_2
\leq C\left( \| X_{j,:} - X_{j^-,:} \|_2 + \sqrt{1 - s_{t,i}} + \gamma M \right),
\]
where $C = L(1+C')$ absorbs constants.
\end{proof}

\end{document}